\pdfoutput=1
\documentclass{article}

\PassOptionsToPackage{numbers, compress}{natbib}
\usepackage[final]{iccv}

\usepackage{times}
\usepackage{epsfig}
\usepackage{graphicx}
\usepackage{amsmath}
\usepackage{amssymb}
\usepackage{times}
\usepackage{epsfig}
\usepackage{graphicx}
\usepackage{amsmath}
\usepackage{amssymb}
\usepackage{multirow}
\usepackage{booktabs} 
\usepackage{algorithm}
\usepackage{algorithmic}
\usepackage{subfigure}

\usepackage{color}

\usepackage[pagebackref=true,breaklinks=true,letterpaper=true,colorlinks,bookmarks=false]{hyperref}

\def\iccvPaperID{157} 

\begin{document}
\title{Generative Semantic Manipulation with Contrasting GAN}

%
\author{Xiaodan Liang, Hao Zhang, Eric P. Xing\\
	Carnegie Mellon University and Petuum Inc. \\
	{\tt\small \{xiaodan1, hao, epxing\}@cs.cmu.edu}
}
\maketitle

\begin{abstract}
  Generative Adversarial Networks (GANs) have recently achieved significant improvement on paired/unpaired image-to-image translation, such as photo$\rightarrow$ sketch and artist painting style transfer. However, existing models can only be capable of transferring the low-level information (e.g. color or texture changes), but fail to edit high-level semantic meanings (e.g., geometric structure or content) of objects. On the other hand, while some researches can synthesize compelling real-world images given a class label or caption, they cannot condition on arbitrary shapes or structures, which largely limits their application scenarios and interpretive capability of model results. In this work, we focus on a more challenging semantic manipulation task, which aims to modify the semantic meaning of an object while preserving its own characteristics (e.g. viewpoints and shapes), such as cow$\rightarrow$sheep, motor$\rightarrow$ bicycle, cat$\rightarrow$dog. To tackle such large semantic changes, we introduce a contrasting GAN (contrast-GAN) with a novel adversarial contrasting objective. Instead of directly making the synthesized samples close to target data  as previous GANs did, our adversarial contrasting objective optimizes over the distance comparisons between samples, that is, enforcing the manipulated data be semantically closer to the real data with target category than the input data. Equipped with the new contrasting objective, a novel mask-conditional contrast-GAN architecture is proposed to enable disentangle image background with object semantic changes. Experiments on several semantic manipulation tasks on ImageNet and MSCOCO dataset show considerable performance gain by our contrast-GAN over other conditional GANs. Quantitative results further demonstrate the superiority of our model on generating manipulated results with high visual fidelity and reasonable object semantics.
\end{abstract}

\section{Introduction}

Arbitrarily manipulating image content given either a target image, class or caption has recently attracted a lot of research interests and would advance a wide range of applications, e.g. image editing and unsupervised representation learning. Recent generative models~\cite{isola2016image,sangkloy2016scribbler,zhu2017unpaired,johnson2016perceptual,yi2017dualgan,li2017perceptual,dai2017scan} have achieved great progresses on modifying low-level content, such as transferring color and texture from a holistic view. However, these models often fail to perform large semantic changes (e.g. cat $\rightarrow$ dog, motor $\rightarrow$ bicycle) which are essential to bridge the gap between high-level concepts and pixel-wise details. 

On the other hand, compelling conditional image synthesis given a specific object category (e.g. ``bird")~\cite{mirza2014conditional,yan2016attribute2image}, a textural description (``a yellow bird with a black head")~\cite{reed2016generative}, or locations~\cite{reed2016learning} has already been demonstrated using variants of Generative Adversarial Networks (GANs)~\cite{goodfellow2014generative,radford2015unsupervised} and Variational Autoencoders~\cite{gregor2015draw}. However, existing approaches have so far only used fixed and simple conditioning variables such as a class or location that can be conveniently formatted as inputs, but failed to control more complex variables (e.g. shapes and viewpoints). It largely limits the application potential of image generation tasks and interpretive capability of unsupervised generative modeling. 

\begin{figure}[!tp]
		\begin{center}
			\includegraphics[scale=0.45]{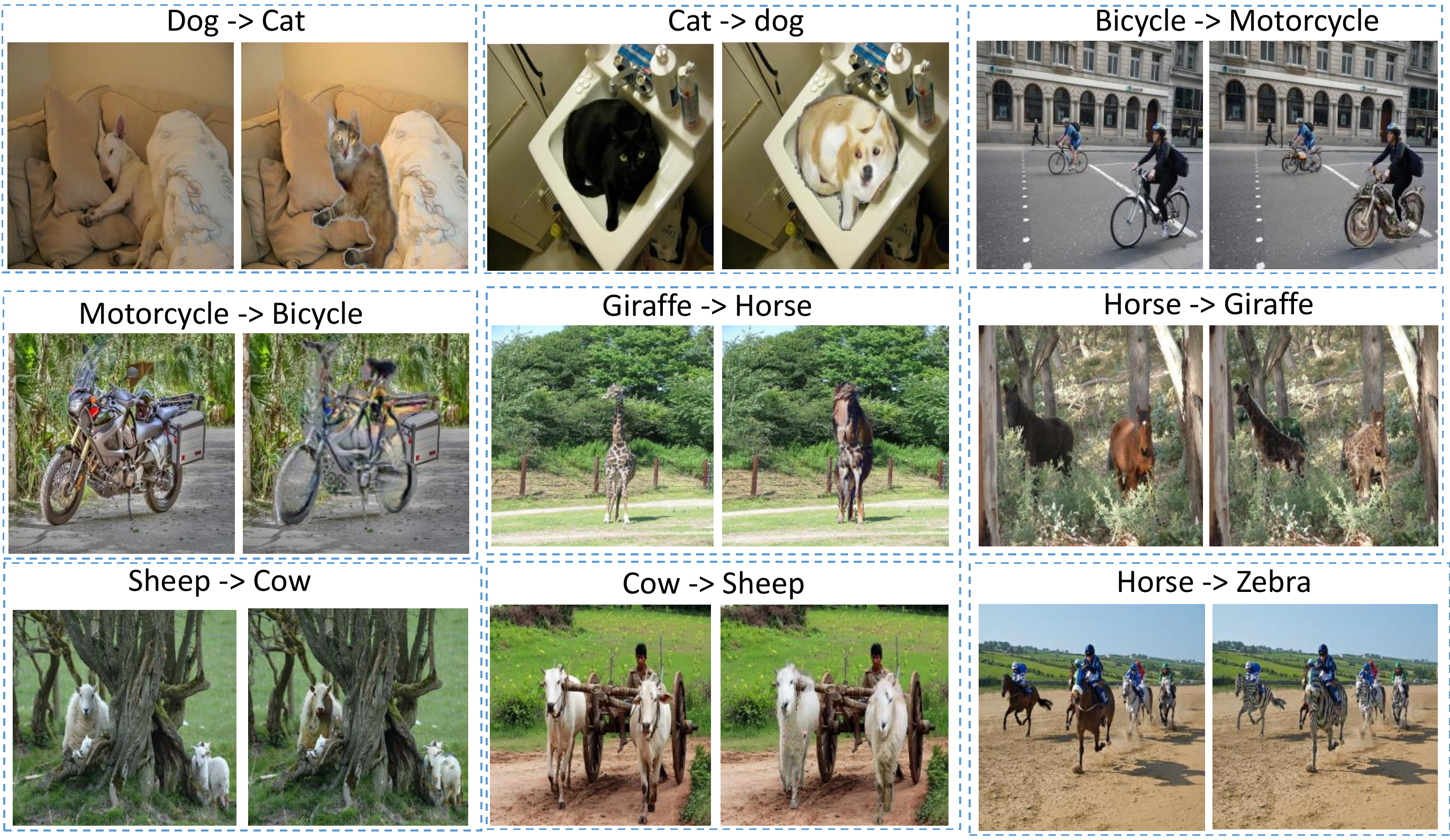}		\vspace{-4mm}
			\caption{Some example semantic manipulation results by our model, which takes one image and a desired object category (e.g. \emph{cat}, \emph{dog}) as inputs and then learns to automatically change the object semantics by modifying their appearance or geometric structure. We show the original image (left) and manipulated result (right) in each pair.} 
			\label{fig:teaser}
		\end{center}
		\vspace{-6mm}
	\end{figure}

In this paper we take a further step towards image semantic manipulation in the absence of any paired training examples. It not only generalizes image-to-image translation research by enabling manipulate high-level object semantics, but also pushes the boundary of controllable image synthesize research by retaining intrinsic characteristics conveyed in the original image as much as possible. Figure~\ref{fig:teaser} shows some example semantic manipulation results by our model. It can be observed that our model tends to perform very few shape, geometric or texture changes over the input image, and yet successfully changes the semantic meaning of the objects into desired ones, such as cat$\rightarrow$dog.

To tackle such large semantic changes, we propose a novel contrasting GAN (contrast-GAN) in the spirit of learning by comparisons~\cite{schroff2015facenet,hoffer2016deep}. Different from the objectives used in previous GANs that often directly compare the target values with the network outputs, the proposed contrast-GAN introduces an adversarial distance comparison objective for optimizing one conditional generator and several semantic-aware discriminators. This contrasting objective enforces that the features of the synthesized samples are much closer to those of real data with target semantic than the input data. Furthermore, in order to disentangle image background from semantic parts, we propose a novel mask-conditional contrast-GAN architecture for realizing the attentive semantic manipulation on the whole image by conditioning on masks of object instances. 

We demonstrate the promising semantic manipulation ability of the proposed contrast-GAN qualtitatively and quantitatively on labels$\leftrightarrow$photos on Cityscape dataset~\cite{cordts2016cityscapes} , apple$\leftrightarrow$orange and horse$\leftrightarrow$zebra on Imagenet~\cite{deng2009imagenet} and ten challenging semantic manipulation tasks (e.g. cat$\leftrightarrow$dog, bicycle$\leftrightarrow$motorcycle) on MSCOCO dataset~\cite{lin2014microsoft}, as illustrated in Figure~\ref{fig:teaser}. We further quantitatively show its superiority compared to existing GAN models~\cite{liu2016coupled,dumoulin2016adversarially,shrivastava2016learning,johnson2016perceptual,zhu2017unpaired} on unpaired image-to-image translation task and more challenging semantic manipulation tasks.

\section{Related Work}

\textbf{Generative Adversarial Networks (GANs).} There has been a large GAN-family methods since the seminal work by Goodfellow et al.~\cite{goodfellow2014generative}. Impressive progresses have been achieved on a wide variety of image generation~\cite{mirza2014conditional,reed2016learning,improvegan}, image editing~\cite{zhu2016generative}, text generation~\cite{liang2017recurrent} and conditional image generation such as text2image~\cite{reed2016generative}, image inpainting~\cite{pathak2016context}, and image translation~\cite{isola2016image} tasks. The key to GANs' success is the variants of adversarial loss that forces the synthesized images to be indistinguishable from real data distribution. To handle the well-known mode collapse issue of GAN and make its training more stable, diverse training objectives have been developed, such as Earth Mover Distance in WGAN~\cite{arjovsky2017wasserstein}, feature matching loss~\cite{improvegan}, loss-sensitive GAN~\cite{qi2017loss}. However, unlike existing GAN objectives that seek an appropriate criterion between synthesized samples and target outputs, we propose a tailored adversarial contrasting objective for image semantic manipulation. Our contrast-GAN is inspired by the strategy of learning by comparison, that is, aiming to learn the mapping function such that the semantic features of manipulated images are much closer to feature distributions of target domain than those of the original domain. 

\begin{figure}[!tp]
		\begin{center}
			\includegraphics[scale=0.65]{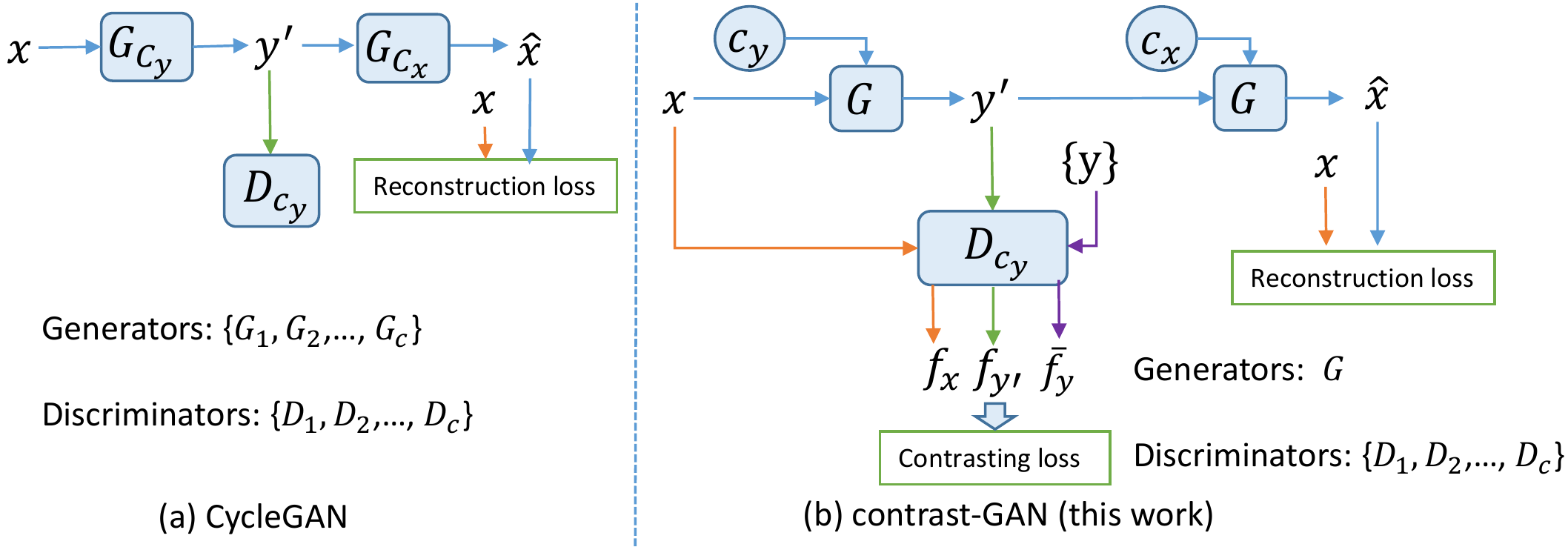}		\vspace{-4mm}
			\caption{An overview of the proposed contrast-GAN. $c_y$ and $c_x$ indicate the object categories (semantics) of domain $X$ and $Y$, respectively. $G_{c_y}$ translates samples into domain $Y$ and $D_{c_y}$ distinguishes between the manipulated result $y'$ and real ones $y$, and vice verse for $G_{c_x}$ and $D_{c_x}$. (a) shows the original CycleGAN in~\cite{zhu2017unpaired} where separate generators and discriminators for each mapping are optimized using the cycle-consistency loss. (b) presents the workflow of our contrast-GAN that optimizes one conditional generator $G$ and several semantic-aware discriminators $D_1, D_2, \dots, D_C$, where $C$ is the total number of object categories. We introduce an adversarial contrasting loss into GAN that encourages the features $f_{y'}$ of generated sample $y'$ are much closer to the feature center $\bar{f}_y$ of target domain $Y$ than those of input $x$.}
			\label{fig:contrastGAN}
		\end{center}
		\vspace{-8mm}
	\end{figure}

\textbf{Generative Image-conditional Models.} GANs have shown great success on a variety of image-conditional models such as style transfer~\cite{johnson2016perceptual,wang2017zm} and general-purpose image-to-image translation~\cite{isola2016image}. More recent approaches~\cite{zhu2017unpaired,yi2017dualgan,liu2017unsupervised,liu2016coupled} have tackled the unpaired setting for cross-domain image translation and also conducted experiments on simple semantic translation (e.g. horse$\rightarrow$zebra and apple$\rightarrow$orange), where only color and texture changes are required. Compared to prior approaches that only transfer low-level information, we focus on high-level semantic manipulation on images given a desired category. The unified mask-controllable contrast-GAN is introduced to disentangle image background with object parts, comprised by one shared conditional generator and several semantic-aware discriminators within an adversarial optimization. Our model can be posed as a general-purpose solution for high-level semantic manipulation, which can facilitate many image understanding task, such as unsupervised/semi-supervised activity recognition and object recognition.

\section{Semantic Manipulation with Contrasting GAN}

The goal of semantic manipulation is to learn mapping functions for manipulating input images into target domains specified by various object semantics $\{c_k\}_{k=1}^{C}$, where $C$ is the total number of target categories. For each semantic $c_k$, we have a set of images $\{{I}_{c_k}\}$. For notation simplicity, we denote the input domain as $X$ with semantic $c_x$ and output domain as 
$Y$ with semantic $c_y$ in each training/testing step. As illustarted in Figure~\ref{fig:contrastGAN}, our contrast-GAN learns a conditional generator $G$, which takes a desired semantic $c_y$ and an input image $x$ as inputs, and then manipulates $x$ into $y'$. The semantic-aware adversarial discriminators $D_{c_y}$ aims to distinguish between images 
$y\in Y$ and manipulated results $y' = G(x, c_y)$. Our new adversarial contrasting loss forces the representations of generated result $y'$ be closer to those of images $\{y\}$ in target domain $Y$ than those of input image $x$.

 In the following sections, we first describe our contrast-GAN architecture and then present the mask-conditional contrast-GAN for disentangling image background and object semantic.
 
 \subsection{Adversarial Contrasting objective}

The adversarial loss introduced in Generative Adversarial Networks (GANs)~\cite{goodfellow2014generative} consists of a generator $G$ and a discriminator $D$ that compete in a two-player min-max game. The objective of vanilla GAN is to make the discriminator correctly classify its inputs as either real or synthetic and the generator synthesize images that the discriminator will classify as real. In practice  we can replace the negative log likelihood objective by a least square loss~\cite{mao2016multi}, which performs more stably during training and generates higher quality results. Thus, the GAN objective becomes:
\begin{equation}
	\begin{split}
	\mathcal{L}_{\text{LSGAN}}(G, D_{c_y}, c_y) = \mathbb{E}_{y \sim p_{\text{data}}(y)}[(D_{c_y}(y)-1)^2] + \mathbb{E}_{x \sim p_{\text{data}}(x)}[D_{c_y}(G(x, c_y))^2].	\label{eq:vanilla}
	\end{split}
	\end{equation}

	\begin{figure}[!tp]
		\begin{center}
			\includegraphics[scale=0.56]{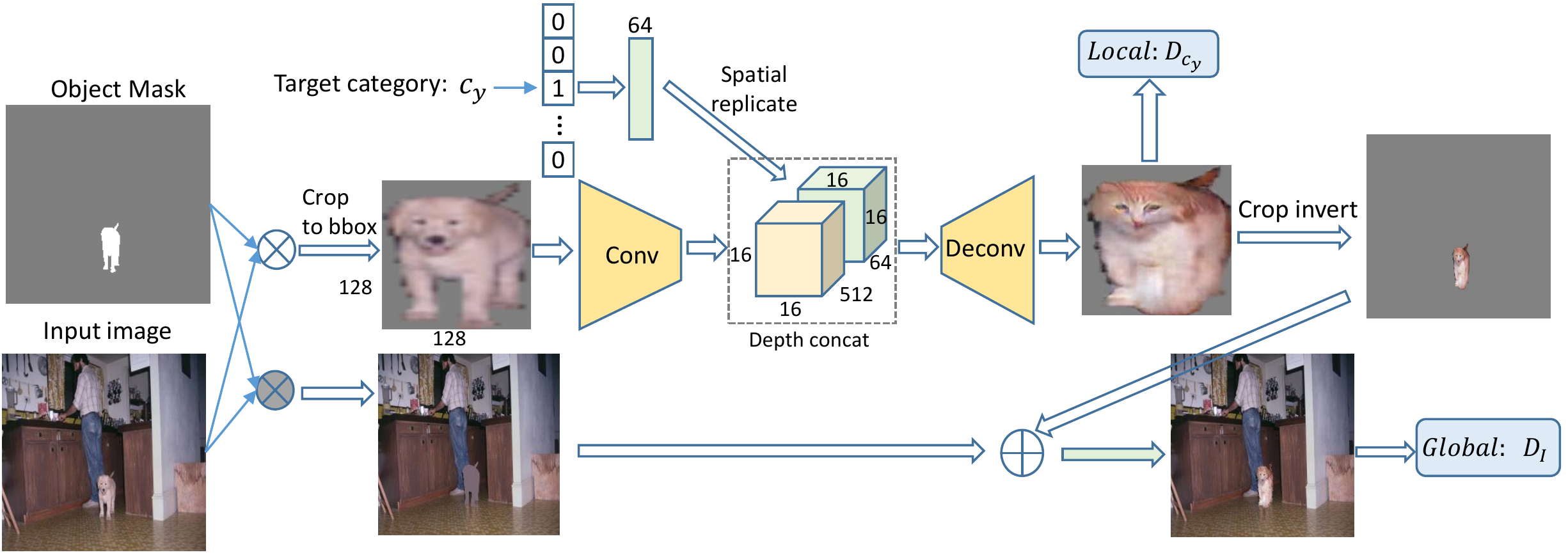}		\vspace{-3mm}
			\caption{The proposed mask-conditional contrast-GAN for semantic manipulation by taking an input image, an object mask and a target category as input. Please refer more details in Section~\ref{sec:mask}. }
			\label{fig:maskGAN}
		\end{center}
		\vspace{-4mm}
	\end{figure}
In this work, in order to tackle large semantic changes, we propose a new adversarial contrasting objective in the spirit of learning by comparison. Using a comparative measure with neural network to learn embedding space was introduced in the ``Siamese network"~\cite{schroff2015facenet,hoffer2016deep} with triple samples. The main idea is to optimize over distance comparisons between generated samples with those from the source domain $X$ and target domain $Y$. We consider the feature representation of manipulated result $y'$ should be closer to those of real data $\{y\}$ in target domain $Y$ than that of $x$ in input domain $X$ under the background of object semantic $c_y$. Formally, we can produce semantic-aware features by feeding the samples into $D_{c_y}$, resulting in $f_{y'}$ for $y'$ served as an anchor sample, $f_{x}$ for the input $x$ as a contrasting sample and $\{f_y\}_N$ for samples $\{y\}_N$ in the target domain as positive samples. Note that, we compare the anchor $f_{y'}$ with the approximate feature center $\bar{f}_y$ computed as the average of all features $\{f_y\}_N$ rather than that of one randomly sampled $y$ in each step, in order to reduce model oscillation. The generator aims to minimize the contrasting distance $Q(\cdot)$:
\begin{equation}
	\begin{split}
	Q(f_{y'}, f_{x}, \bar{f}_y) = -\log\frac{e^{-||f_{y'} - \bar{f}_y||_2}}{e^{-||f_{y'} - \bar{f}_y||_2} + e^{-||f_{y'} - f_x||_2}} .
	\end{split}
	\label{eq:featurecenter}
	\end{equation}
Similar to the target of $D_{c_y(y)}$ in Eq.(\ref{eq:vanilla}) that tries to correctly classify its inputs as either real or fake, our discriminator aims to maximize the contrasting distance $Q(f_{y'}, f_{x}, \bar{f}_y)$. The adversarial contrasting objective for GAN can be defined as:
\begin{equation}
	\begin{split}
	\mathcal{L}_{\text{contrast}}(G, D_{c_y}, c_y) =
	\mathbb{E}_{y \sim p_{\text{data}}(y), x \sim p_{\text{data}(x)}}[Q(D_{c_y}(G(x, c_y)), D_{c_y}(x), D_{c_y}(\{y\}))].
	\label{eq:contrast}
	\end{split}
	\end{equation}

To further reduce the space of possible mapping functions by the conditional generator, we also use the cycle-consistency loss in~\cite{zhu2017unpaired} which constrains the mappings (induced by the generator $G$) between two object semantics should be inverses of each other. Notably, different from ~\cite{zhu2017unpaired} which used independent generators for each domain, we use a single shared conditional generator for all domains. The cycle objective can be defined as: 
\begin{equation}
	\begin{split}
	\mathcal{L}_{\text{cycle}}(G, c_y, c_x) =
	\mathbb{E}_{x \sim p_{\text{data}(x)}}[||G(G(x, c_y), c_x) - x ||_1].
	\label{eq:cycle}
	\end{split}
	\end{equation}

Therefore, our full objective is computed by combining Eq.(\ref{eq:vanilla}), Eq.(\ref{eq:contrast}) and Eq.(\ref{eq:cycle}):
\begin{equation}
	\begin{split}
	\mathcal{L}_{\text{contrast-GAN}}(G, D_{c_y}, c_y) =\mathcal{L}_{\text{contrast}}(G, D_{c_y}, c_y) + \lambda\mathcal{L}_{\text{LSGAN}}(G, D_{c_y}, c_y)  + \beta \mathcal{L}_{\text{cycle}}(G, c_y, c_x),
	\label{eq:full}
	\end{split}
	\end{equation}
where $\lambda$ and $\beta$ control the relative importance of the objectives. $G$ tries to minimize this objective against a set of  adversarial discriminators $\{D_{c_y}\}$ that tries to maximize them, i.e. $G^* = \arg\min_G(\frac{1}{C}\sum_{c_y}\max_{D_{c_y}} \mathcal{L}_{\text{contrast-GAN}}(G,D_{c_y},c_y)$). Our extensive experiments show that each of objectives plays a critical role in arriving at high-quality manipulation results.

\subsection{Mask-conditional Contrast-GAN}
\vspace{-2mm}
\label{sec:mask}
Figure~\ref{fig:maskGAN} shows a sketch of our model, which starts from an input image $x$, an object mask $M$ and target category $c_y$ and outputs the manipulated image. Note that the whole architecture is fully differential for back-propagation. For clarity, the full cycle architecture (i.e. the mapping $y'\rightarrow \hat{x}$ via $G(y, c_x)$) is omitted in Figure~\ref{fig:maskGAN}. Below we walk through each step.

\begin{figure}[!tp]
		\begin{center}
			\includegraphics[scale=0.35]{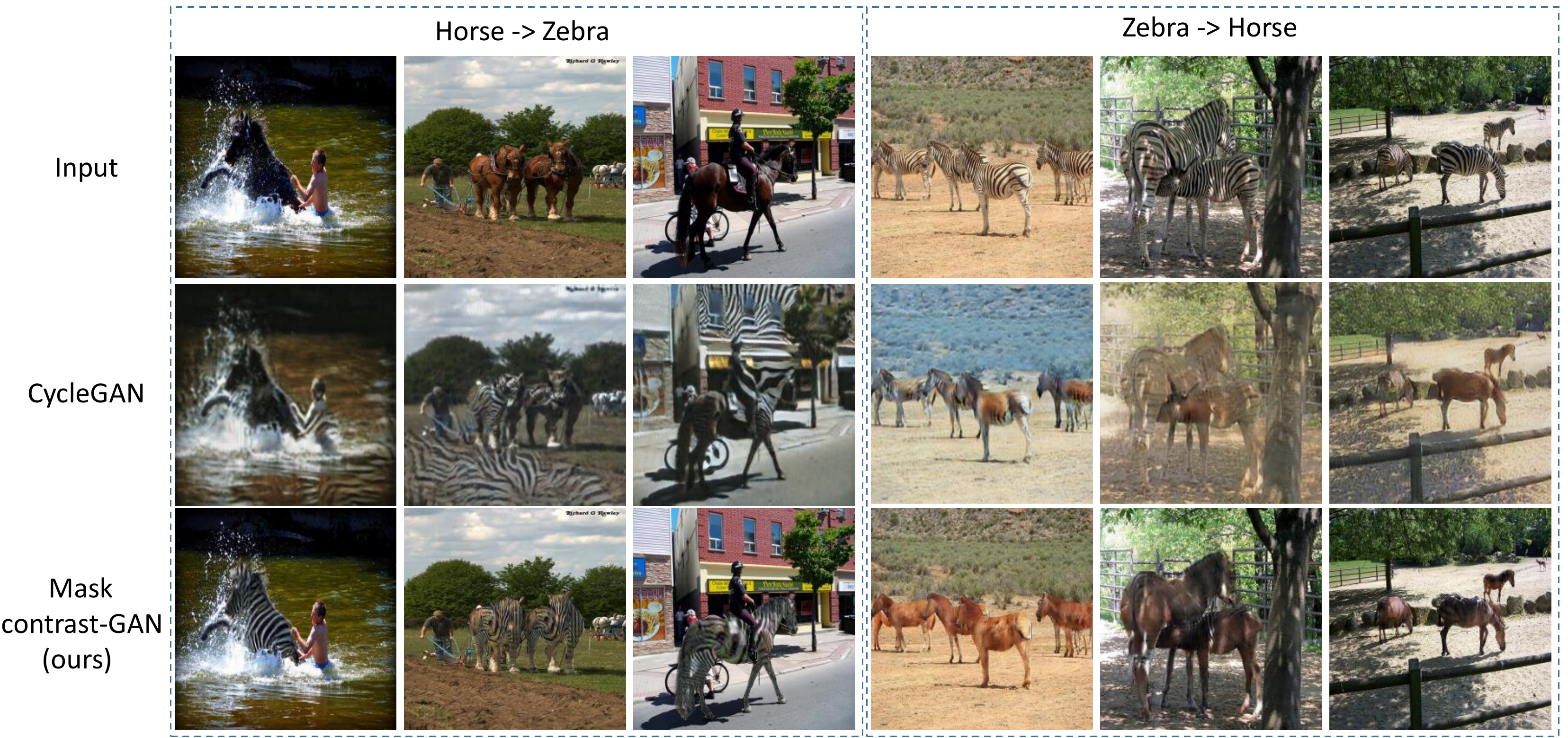}		\vspace{-2mm}
			\caption{Result comparisons between our mask contrast-GAN with CycleGAN~\cite{zhu2017unpaired} for translating horse$\rightarrow$zebra and zebra$\rightarrow$horse on the MSCOCO dataset with provided object masks. It shows the effectiveness of incorporating object masks to disentangle image background and object semantics.}  
			\label{fig:compare_mask}
		\end{center}
		\vspace{-8mm}
	\end{figure}
First, a masking operation and subsequent spatial cropping operation are performed to obtain the object region with size of $128\times128$. The background image is calculated by functioning the inverse mask map on input image. The object region is then fed into several convolutional layers to get $16\times16$ feature maps with $512$ dimension. Second, we represent the target category $c_y$ using an one-hot vector which is then passed into a linear layer to get a feature embedding with $64$ dimension. This feature is replicated spatially to form a $16\times 16 \times 64$ feature maps, and then concatenated with image feature maps via the depth concatenation. Third, several deconvolution layers are employed to obtain target region with $128\times128$. We then warp the manipulated region back into the original image resolution, which is then combined with the background image via an additive operation to get the final manipulated image. We implement the spatial masking and cropping modules using spatial transformers~\cite{jaderberg2015spatial}.

To enforce the semantic manipulation results be semantically consistent with both the target semantic and the background appearance of input image, we adopt both local discriminators $\{D_{c_y}\}$ defined in our contrast-GAN and a global image discriminator $D_I$. Each local discriminator $D_{c_y}$ is responsible for verifying whether the high-level semantic of outputs is semantically coherent with the input target while the global one $D_I$ evaluates the visual fidelity of the whole manipulated image.
\subsection{Implementation Details}
\vspace{-2mm}
\textbf{Network Architecture.} To make a fair comparison, We adopt similar architectures from~\cite{zhu2017unpaired} which have shown impressive results for unpaired image translation. This generator contains three stride-2 convolutions, six residual blocks, and three fractionally strided convolutions. For the architecture of mask-conditional contrast-GAN in Figure~\ref{fig:maskGAN}, the residual blocks are employed after concatenating convolutional feature maps with maps of the target category. In terms of the target category input for generator $G$, we specify different number of categories $C$ for each dataset, such as $C=10$ for ten semantic manipulation tasks on MSCOCO dataset. We use the same patch-level discriminator used in~\cite{zhu2017unpaired} for local discriminators $\{D_{c_y}\}$ and the global discriminator $D_I$.   

\textbf{Training Details.} To compute the approximate feature center $\bar{f}_y$ in Eq.(\ref{eq:featurecenter}) for the contrasting objective, we keep an image buffer with randomly selected $N=50$ samples in target domain $Y$. For all the experiments, we set $\lambda= 10$ and $\beta=10$ in Eq.(\ref{eq:full}) to balance each objective. We use the Adam solver~\cite{kingma2014adam} with a batch size of 1. All networks
were trained from scratch, and trained with learning rate of 0.0002 for the first 100 epochs and a linearly decaying rate that goes to zero over the next 100 epochs. Our algorithm only optimizes over one conditional generator and several semantic-aware discriminators for all kinds of object semantics. All models are implemented on Torch framework.

\begin{table}[!tp]\setlength{\tabcolsep}{1pt}
	\centering\scriptsize
	\parbox{.45\linewidth}{\caption{Comparison of FCN-scores on Cityscapes labels$\rightarrow$photos dataset.}\label{tab:label2photo}
	\begin{tabular}{cccccccccccccccccccccc}
		\toprule
		{Method} &  Per-pixel acc. &  Per-class acc.  &  Class IOU \\
		\midrule
		CoGAN~\cite{liu2016coupled}  & 0.40 & 0.10 & 0.06 \\
		BiGAN~\cite{dumoulin2016adversarially}  & 0.19 & 0.06 & 0.02\\
		Pixel loss+ GAN~\cite{shrivastava2016learning} & 0.20 & 0.10 & 0.0\\
		Feature loss+GAN~\cite{johnson2016perceptual} & 0.07 & 0.04 & 0.01\\
		CycleGAN~\cite{zhu2017unpaired} & 0.52 & 0.17 & 0.11 \\
		\hline
		{Contrast alone} & {0.53} & {0.13} & {0.12} \\
		Contrast + classify & 0.55 & 0.15 & 0.11 \\
		Contrast + Cycle & 0.57 & 0.22 & 0.13\\
		Contrast-GAN (separate G) & 0.57 & \textbf{0.22} & \textbf{0.17}\\
		\textbf{Contrast-GAN (ours)} & \textbf{0.58} & {0.21} & {0.16} \\
		\hline
				\vspace{-7mm}
	\end{tabular}%
	}
	\hfill
	\parbox{.45\linewidth}{\caption{Comparison of classification performance on Cityscapes photos$\rightarrow$ labels dataset. }\label{tab:photo2label}
	\begin{tabular}{cccccccccccccccccccccc}
		\toprule
		{Method} &  Per-pixel acc. &  Per-class acc.  &  Class IOU \\
		\midrule
		CoGAN~\cite{liu2016coupled}  & 0.45 & 0.11 & 0.08 \\
		BiGAN~\cite{dumoulin2016adversarially}  & 0.41 & 0.13 & 0.07\\
		Pixel loss+ GAN~\cite{shrivastava2016learning} & 0.47 & 0.11 & 0.07\\
		Feature loss+GAN~\cite{johnson2016perceptual} & 0.50 & 0.10 & 0.06\\
		CycleGAN~\cite{zhu2017unpaired} & 0.58 & 0.22 & 0.16 \\
		\hline
		{Contrast alone} & {0.54} & {0.12} & {0.10} \\
		Contrast + classify & 0.55 & 0.13 & 0.11 \\
	    Contrast + Cycle & 0.60 & 0.19 & 0.15\\
	    Contrast-GAN (separate G) & 0.60 & 0.23 & 0.17\\
		\textbf{Contrast-GAN (ours)} & \textbf{0.61} & \textbf{0.23} & \textbf{0.18} \\
		\hline
				\vspace{-6mm}
	\end{tabular}%
	}
\end{table}%
\begin{figure}[!tp]
		\begin{center}
			\includegraphics[scale=0.35]{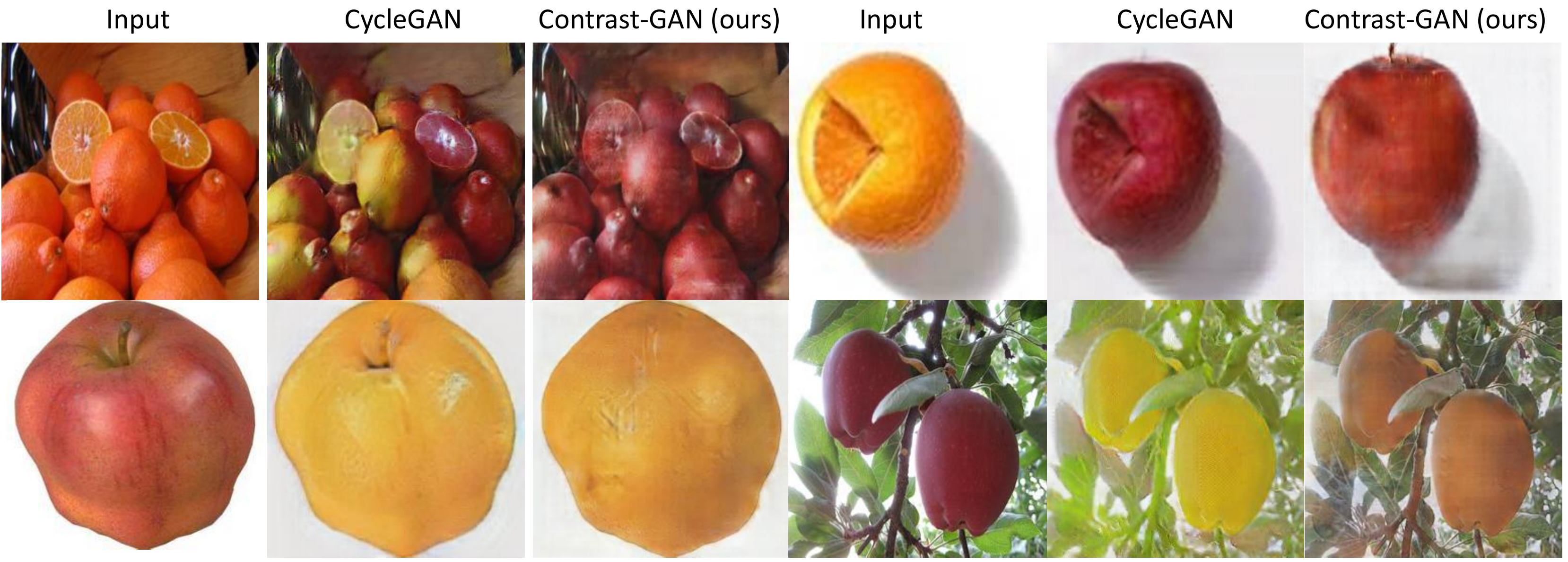}		\vspace{-2mm}
			\caption{Result comparison by our contrast-GAN with CycleGAN~\cite{zhu2017unpaired} for translating orange$\rightarrow$apple (first row) and apple$\rightarrow$orange (second row) on ImageNet.
			}
			\label{fig:apple}
		\end{center}
		\vspace{-6mm}
	\end{figure}
	
\section{Experiments}

\textbf{Datasets.} First, we quantitatively compare the proposed contrast-GAN against recent state-of-the-arts on the task of labels$\leftrightarrow$photos on the Cityscape dataset~\cite{cordts2016cityscapes}. The labels$\leftrightarrow$Photos dataset uses images from Cityscape training set for training and val set for testing. Following~\cite{zhu2017unpaired}, we use the unpaired setting during training and the ground truth input-output pairs for evaluation. Second, we compare our contrast-GAN with CycleGAN~\cite{zhu2017unpaired} on unpaired translation, evaluating on the task of horse$\leftrightarrow$zebra and apple$\leftrightarrow$orange from ImageNet. The images for each class are downloaded from ImageNet~\cite{deng2009imagenet} and scaled to 128$\times$128, consisting of 939 images for horse, 1177 for zebra, 996 for apple and 1020 for orange. Finally, we apply contrast-GAN into ten more challenging semantic manipulation tasks, i.e. dog$\leftrightarrow$cat, cow$\leftrightarrow$sheep, bicycle$\leftrightarrow$motorcycle, horse$\leftrightarrow$giraffe, horse$\leftrightarrow$zebra. To disentangle image background with the object semantic information, we test the performance of mask-conditional architecture. The mask annotations for each image are obtained from MSCOCO dataset~\cite{lin2014microsoft}. For each object category, the images in MSCOCO train set are used for training and those in MSCOCO val set for testing. The output realism of manipulated results by different methods is quantitatively compared by AMT perception studies described below.

\textbf{Evaluation Metrics.} We adopt the ``FCN score" from~\cite{isola2016image} to evaluate Cityscapes labels$\rightarrow$photo task, which evaluates how interpretable the generated photos are according to an off-the-shelf semantic segmentation algorithm. To evaluate the performance
of photo$\rightarrow$labels, we use the standard ``semantic segmentation metrics" from Cityscapes benchmark, including per-pixel accuracy,
per-class accuracy, and mean class Intersection-Over-Union~\cite{cordts2016cityscapes}. For semantic manipulation tasks on ImageNet and MSCOCO datasets (e.g. cat$\rightarrow$dog), we run real vs.fake AMT perceptual studies to compare the realism of outputs from different methods under the background of a specific object semantic (e.g. dog), similar to~\cite{zhu2017unpaired}. For each semantic manipulation task, we collect 10 annotations for randomly selected 100 manipulated images by each method. 

\subsection{Result Comparisons}

\textbf{Labels$\leftrightarrow$photos on Cityscape.} Table~\ref{tab:label2photo} and Table~\ref{tab:photo2label} report the performance comparison on the labels$\rightarrow$photos task and photos$\rightarrow$labels task on Cityscape, respectively. In both cases, the proposed contrast-GAN with a new adversarial contrasting objective outperforms the state-of-the-arts~\cite{liu2016coupled,dumoulin2016adversarially,shrivastava2016learning,johnson2016perceptual,zhu2017unpaired} on unpaired image-to-image translation. Note that we adopt the same baselines~\cite{liu2016coupled,dumoulin2016adversarially,shrivastava2016learning,johnson2016perceptual} for fair comparison in~\cite{zhu2017unpaired}.

	\begin{figure}[!tp]
		\begin{center}
			\includegraphics[scale=0.44]{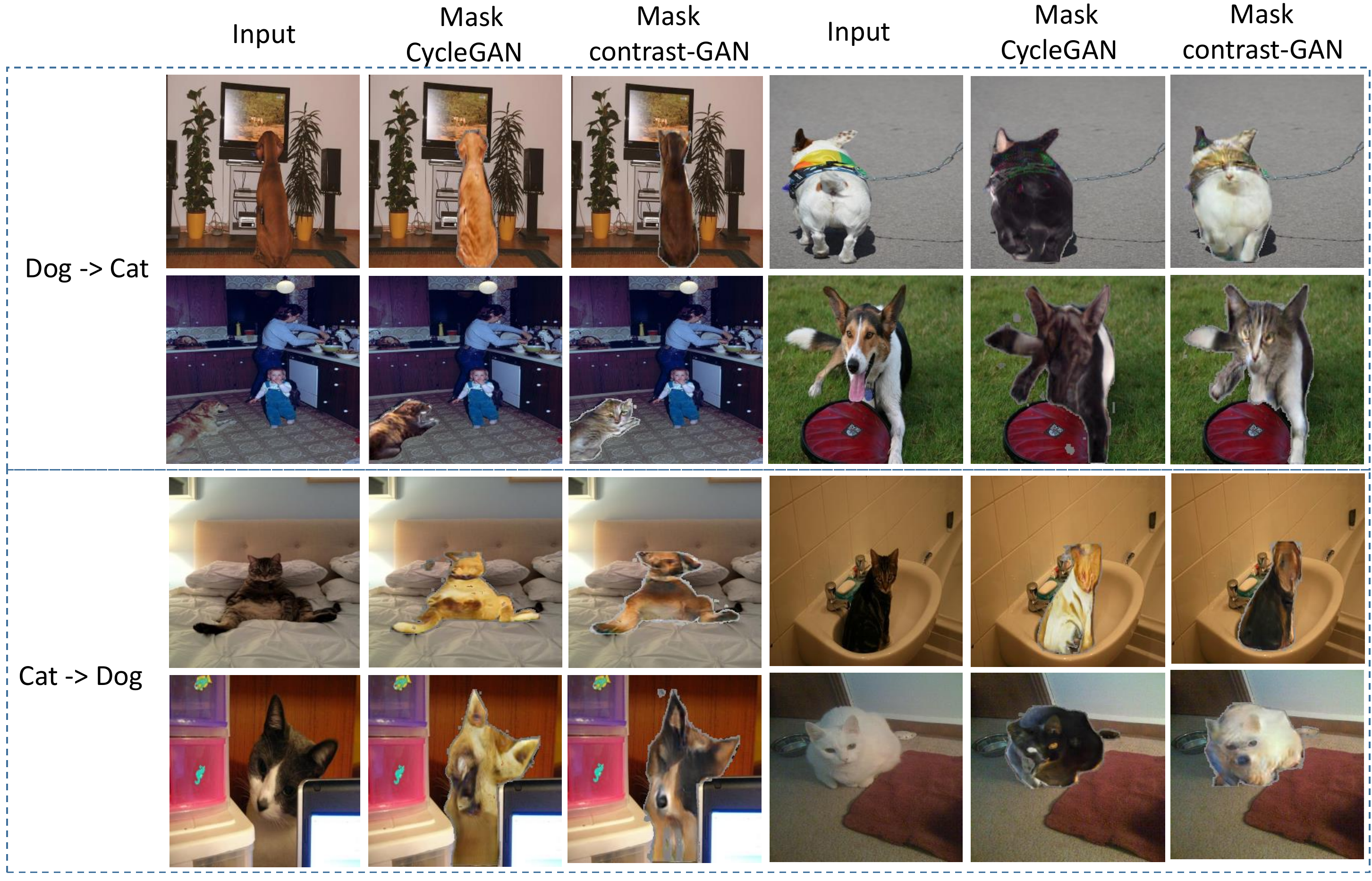}		\vspace{-4mm}
			\caption{Result comparison between our mask contrast-GAN with mask CycleGAN~\cite{zhu2017unpaired} for translating dog$\rightarrow$cat and cat$\rightarrow$dog on the MSCOCO dataset with provided object masks. 
			} 
			\label{fig:compare_mask_cycle}
		\end{center}
		\vspace{-5mm}
	\end{figure}

\begin{table}[!tp]\setlength{\tabcolsep}{1pt}
	\centering\scriptsize
	\caption{Result comparison of AMT perception test on  eight mask-conditional semantic-manipulation tasks on the MSCOCO dataset. The numbers indicate \% images that Turkers labeled real.}\label{tab:amt}\vspace{-2mm}
	\begin{tabular}{cccccccccccccccccccccc}
		\toprule
		{Method} &  cat$\rightarrow$dog &  dog$\rightarrow$cat  & bicycle$\rightarrow$motor & motor$\rightarrow$bicycle & horse$\rightarrow$ giraffe & giraffe$\rightarrow$ horse & cow$\rightarrow$sheep & sheep$\rightarrow$cow\\
		\hline
		Mask CycleGAN & 2.5\% & 4.1\% & 10.9\% & 15.6\% & 1.5\% & 2.3\% & 16.3\% & 18.9\%\\
		Mask Contrast alone & 3.7\% & 5.0\% & 9.3\% & 13.1\% & 1.6\% & 1.8\% & 17.1\% & 15.5\%\\
		Mask Contrast-GAN w/o $D_I$ & 4.3\% & 6.0\% & 12.8\% & 15.7\% & 1.9\% & 4.5\% & 18.3\% & 19.1\%\\
		\textbf{Mask Contrast-GAN (ours)} & \textbf{4.8\%} & \textbf{6.2\%} & \textbf{13.0\%} & \textbf{16.7\%} & \textbf{1.9\%} & \textbf{5.4\%} & \textbf{18.7\%} & \textbf{20.5\%}\\
		\hline
				\vspace{-8mm}
	\end{tabular}
\end{table}%
\textbf{Apple$\leftrightarrow$orange and horse$\leftrightarrow$zebra on ImageNet.} Figure~\ref{fig:apple} shows some example results by the baseline CycleGAN~\cite{zhu2017unpaired} and our contrast-GAN on the apple$\leftrightarrow$orange semantic manipulation. It can be seen that our method successfully transforms the semantic of objects while CycleGAN only tends to modify low-level characteristics (e.g. color and texture). We also perform real vs. fake AMT perceptual studies on both apple$\leftrightarrow$orange and horse$\leftrightarrow$zebra tasks. Our contrast-GAN can fool participants much better than CycleGAN~\cite{zhu2017unpaired} by comparing the number of manipulated images that Turkers labeled real, that is $14.3\%$ vs $12.8\%$ on average for apple$\leftrightarrow$orange and $10.9\%$ vs $9.6\%$ on average for horse$\leftrightarrow$zebra.  

\textbf{Semantic manipulation tasks on MSCOCO.} We further demonstrate the effectiveness of our method on ten challenging semantic manipulation applications with large semantic changes. Figure~\ref{fig:compare_mask} compares the results by our mask-conditional contrast-GAN against CycleGAN~\cite{zhu2017unpaired} on horse$\leftrightarrow$zebra task. It can be seen that CycleGAN often renders the whole image with the target texture and ignores the particular image content at different locations. For instance, CycleGAN wrongly translates the unrelated objects (e.g. person, building) and background as the stripe texture in the horse$\rightarrow$zebra case. On the contrary, our mask contrast-GAN shows appealing results by selectively manipulating objects of interest (e.g. horse) into the desired semantic (e.g. zebra), benefiting from our mask-conditional architecture and adversarial contrasting objective.

Figure~\ref{fig:compare_mask_cycle} visualizes the comparisons of our mask-conditional architecture using cycle-consistency loss in~\cite{zhu2017unpaired} and our contrasting objective, that is, mask CycleGAN vs mask contrast-GAN. The baseline method often tries to translate very low-level information (e.g. color changes) and fail to edit the shapes and key characteristic (e.g. structure) that truly convey a specific high-level object semantic. However, our contrast-GAN tends to perform trivial yet critical changes on objects to satisfy the target semantic while preserving the original object characteristics. In table~\ref{tab:amt}, we report quantitative results on the AMT perceptual realism measure for eight semantic manipulation tasks. It can be observed that our method substantially outperforms the baseline on all tasks, especially on those requiring large semantic changes (e.g. cat$\leftrightarrow$dog and bicycle$\leftrightarrow$motorcycle). In Figure~\ref{fig:results}, we show more qualitative results. Our model shows the promising capability of manipulating object semantics while retaining original shapes, viewpoints and interactions with background.

	\begin{figure}[!tp]
		\begin{center}
			\includegraphics[scale=0.45]{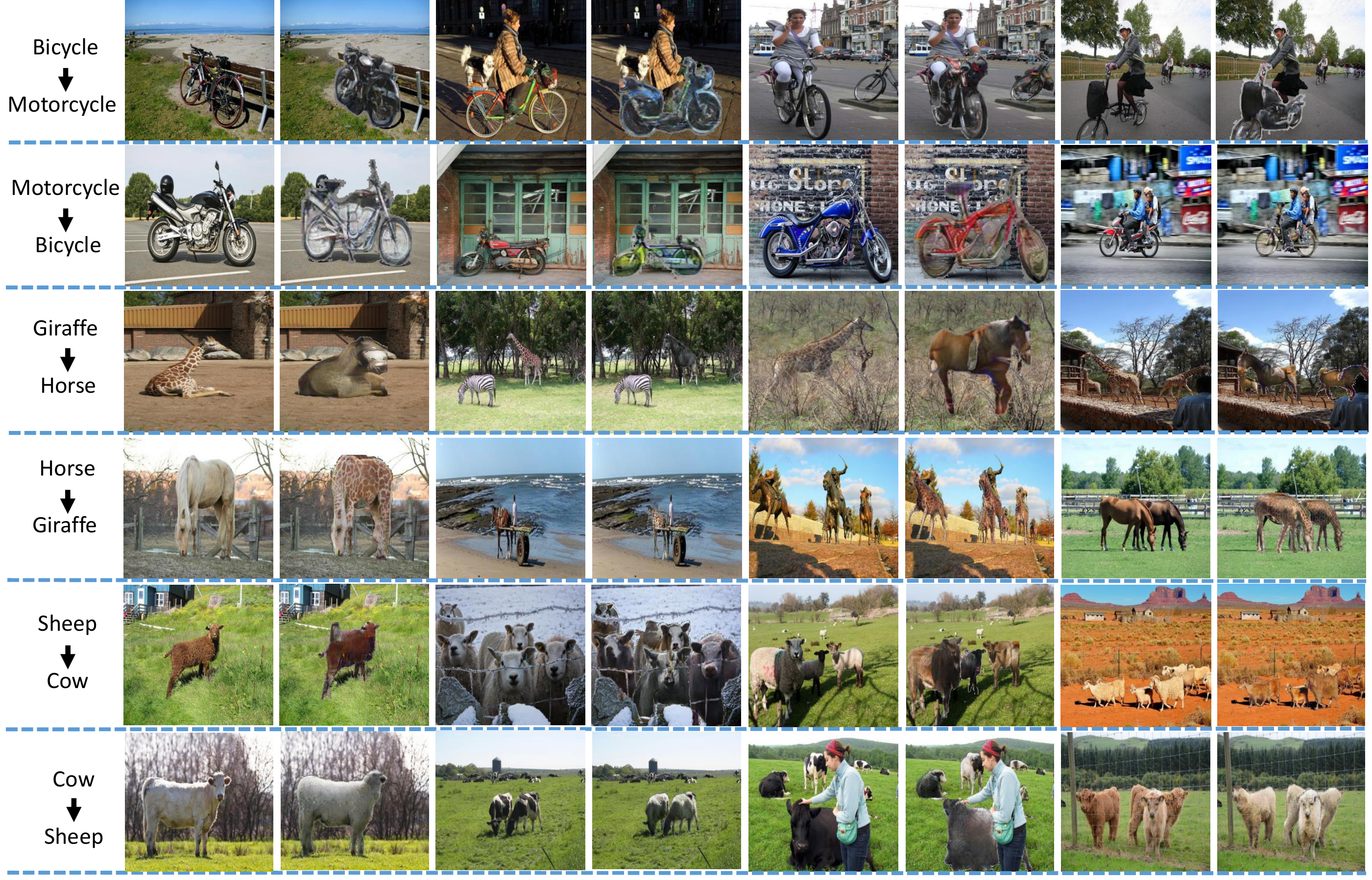}		\vspace{-4mm}
			\caption{Example results by our mask contrast-GAN for manipulating a variety of object semantics on MSCOCO dataset. For each image pair, we show the original image (left) and manipulated image (right) by specifying a  desirable object semantic.} 
			\label{fig:results}
		\end{center}
		\vspace{-6mm}
	\end{figure}
	
\subsection{Model Analysis}
\vspace{-3mm}

In Table~\ref{tab:label2photo} and Table~\ref{tab:photo2label}, we report the results by different variants of our full model on Cityscape labels$\leftrightarrow$photos task. ``Contrast alone" indicates the model only uses $\mathcal{L}_{\text{contrast}}$ as the final objective in Eq.(\ref{eq:full}) while ``Contrast + classifiy" represents the usage of combining of $\mathcal{L}_{\text{contrast}}$ and $\mathcal{L}_{\text{LSGAN}}$ as the final objective. ``Contrast + cycle" is the variant that removes $\mathcal{L}_{\text{LSGAN}}$. CycleGAN~\cite{zhu2017unpaired} can also be regarded as one simplified version of our model that removes the contrasting objective. Table~\ref{tab:amt} shows the ablation studies on mask-conditional semantic manipulation tasks on MSCOCO dataset. It can be seen that ``Contrast alone" and ``Mask Contrast alone" achieve comparable results with the state-of-the-arts. Removing the original \emph{classification}-like objective  $\mathcal{L}_{\text{LSGAN}}$ degrades results compared to our full model, as does removing the cycle-consistency objective $\mathcal{L}_{\text{Cycle}}$. Therefore, we can conclude that all three objectives are critical for performing the semantic manipulation. $\mathcal{L}_{\text{LSGAN}}$ can be complementary with our contrasting objective $\mathcal{L}_{\text{contrast}}$ on validating the visual fidelity of manipulated results. We also validate the advantage of using an auxiliary global discriminator $D_I$ by comparing ``Mask Contrast-GAN w/o $D_I$" and our full model in Table~\ref{tab:amt}.

Note that instead of using separate generators for each semantic as in previous works~\cite{liu2016coupled,dumoulin2016adversarially,shrivastava2016learning,johnson2016perceptual,zhu2017unpaired}, we propose to employ a conditional generator shared for all object semantics. Using one conditional generator has two advantages: first, it can lead to more powerful and robust feature representation by learning over more diverse samples of different semantics; second, the model size can be effectively reduced by only feeding different target categories as inputs to achieve different semantic manipulations. Table~\ref{tab:label2photo} and Table~\ref{tab:photo2label} also report the results of using separate generators for each semantic task in our model, that is, ``Contrast-GAN (separate G)". We can see that our full model using only one conditional generator shows slightly better results than ``Contrast-GAN (separate G)".

\section{Discussion and Future Work}
\vspace{-3mm}
Although our method can achieve compelling results in many semantic manipulation tasks, it shows little success for some cases which require very large geometric changes, such as car$\leftrightarrow$truck and car$\leftrightarrow$bus. Integrating spatial transformation layers for explicitly learning pixel-wise offsets may help resolve very large geometric changes. To be more general, our model can be extended to replace the mask annotations with the predicted object masks or automatically learned attentive regions via attention modeling. This paper pushes forward the research of unsupervised setting by demonstrating the possibility of manipulating high-level object semantics rather than the low-level color and texture changes as previous works did. In addition, it would be more interesting to develop techniques that are able to manipulate object interactions and activities in images/videos for the future work.

{\small
	\bibliographystyle{ieee}
	\bibliography{egbib}

\begin{thebibliography}{10}\itemsep=-1pt

\bibitem{arjovsky2017wasserstein}
M.~Arjovsky, S.~Chintala, and L.~Bottou.
\newblock Wasserstein gan.
\newblock {\em arXiv preprint arXiv:1701.07875}, 2017.

\bibitem{cordts2016cityscapes}
M.~Cordts, M.~Omran, S.~Ramos, T.~Rehfeld, M.~Enzweiler, R.~Benenson,
  U.~Franke, S.~Roth, and B.~Schiele.
\newblock The cityscapes dataset for semantic urban scene understanding.
\newblock In {\em CVPR}, pages 3213--3223, 2016.

\bibitem{dai2017scan}
W.~Dai, J.~Doyle, X.~Liang, H.~Zhang, N.~Dong, Y.~Li, and E.~P. Xing.
\newblock Scan: Structure correcting adversarial network for chest x-rays organ
  segmentation.
\newblock {\em arXiv preprint arXiv:1703.08770}, 2017.

\bibitem{deng2009imagenet}
J.~Deng, W.~Dong, R.~Socher, L.-J. Li, K.~Li, and L.~Fei-Fei.
\newblock Imagenet: A large-scale hierarchical image database.
\newblock In {\em CVPR}, pages 248--255, 2009.

\bibitem{dumoulin2016adversarially}
V.~Dumoulin, I.~Belghazi, B.~Poole, A.~Lamb, M.~Arjovsky, O.~Mastropietro, and
  A.~Courville.
\newblock Adversarially learned inference.
\newblock {\em arXiv preprint arXiv:1606.00704}, 2016.

\bibitem{goodfellow2014generative}
I.~Goodfellow, J.~Pouget-Abadie, M.~Mirza, B.~Xu, D.~Warde-Farley, S.~Ozair,
  A.~Courville, and Y.~Bengio.
\newblock Generative adversarial nets.
\newblock In {\em NIPS}, pages 2672--2680, 2014.

\bibitem{gregor2015draw}
K.~Gregor, I.~Danihelka, A.~Graves, D.~J. Rezende, and D.~Wierstra.
\newblock Draw: A recurrent neural network for image generation.
\newblock {\em arXiv preprint arXiv:1502.04623}, 2015.

\bibitem{hoffer2016deep}
E.~Hoffer, I.~Hubara, and N.~Ailon.
\newblock Deep unsupervised learning through spatial contrasting.
\newblock {\em arXiv preprint arXiv:1610.00243}, 2016.

\bibitem{isola2016image}
P.~Isola, J.-Y. Zhu, T.~Zhou, and A.~A. Efros.
\newblock Image-to-image translation with conditional adversarial networks.
\newblock {\em arXiv preprint arXiv:1611.07004}, 2016.

\bibitem{jaderberg2015spatial}
M.~Jaderberg, K.~Simonyan, A.~Zisserman, et~al.
\newblock Spatial transformer networks.
\newblock In {\em NIPS}, pages 2017--2025, 2015.

\bibitem{johnson2016perceptual}
J.~Johnson, A.~Alahi, and L.~Fei-Fei.
\newblock Perceptual losses for real-time style transfer and super-resolution.
\newblock In {\em ECCV}, pages 694--711, 2016.

\bibitem{kingma2014adam}
D.~Kingma and J.~Ba.
\newblock Adam: A method for stochastic optimization.
\newblock {\em arXiv preprint arXiv:1412.6980}, 2014.

\bibitem{li2017perceptual}
J.~Li, X.~Liang, Y.~Wei, T.~Xu, J.~Feng, and S.~Yan.
\newblock Perceptual generative adversarial networks for small object
  detection.
\newblock {\em arXiv preprint arXiv:1706.05274}, 2017.

\bibitem{liang2017recurrent}
X.~Liang, Z.~Hu, H.~Zhang, C.~Gan, and E.~P. Xing.
\newblock Recurrent topic-transition gan for visual paragraph generation.
\newblock {\em arXiv preprint arXiv:1703.07022}, 2017.

\bibitem{lin2014microsoft}
T.-Y. Lin, M.~Maire, S.~Belongie, J.~Hays, P.~Perona, D.~Ramanan,
  P.~Doll{\'a}r, and C.~L. Zitnick.
\newblock Microsoft coco: Common objects in context.
\newblock In {\em ECCV}, pages 740--755, 2014.

\bibitem{liu2017unsupervised}
M.-Y. Liu, T.~Breuel, and J.~Kautz.
\newblock Unsupervised image-to-image translation networks.
\newblock {\em arXiv preprint arXiv:1703.00848}, 2017.

\bibitem{liu2016coupled}
M.-Y. Liu and O.~Tuzel.
\newblock Coupled generative adversarial networks.
\newblock In {\em NIPS}, pages 469--477, 2016.

\bibitem{mao2016multi}
X.~Mao, Q.~Li, H.~Xie, R.~Y. Lau, and Z.~Wang.
\newblock Multi-class generative adversarial networks with the l2 loss
  function.
\newblock {\em arXiv preprint arXiv:1611.04076}, 2016.

\bibitem{mirza2014conditional}
M.~Mirza and S.~Osindero.
\newblock Conditional generative adversarial nets.
\newblock {\em arXiv preprint arXiv:1411.1784}, 2014.

\bibitem{pathak2016context}
D.~Pathak, P.~Krahenbuhl, J.~Donahue, T.~Darrell, and A.~A. Efros.
\newblock Context encoders: Feature learning by inpainting.
\newblock In {\em CVPR}, pages 2536--2544, 2016.

\bibitem{qi2017loss}
G.-J. Qi.
\newblock Loss-sensitive generative adversarial networks on lipschitz
  densities.
\newblock {\em arXiv preprint arXiv:1701.06264}, 2017.

\bibitem{radford2015unsupervised}
A.~Radford, L.~Metz, and S.~Chintala.
\newblock Unsupervised representation learning with deep convolutional
  generative adversarial networks.
\newblock {\em arXiv preprint arXiv:1511.06434}, 2015.

\bibitem{reed2016generative}
S.~Reed, Z.~Akata, X.~Yan, L.~Logeswaran, B.~Schiele, and H.~Lee.
\newblock Generative adversarial text to image synthesis.
\newblock In {\em ICML}, 2016.

\bibitem{reed2016learning}
S.~E. Reed, Z.~Akata, S.~Mohan, S.~Tenka, B.~Schiele, and H.~Lee.
\newblock Learning what and where to draw.
\newblock In {\em NIPS}, pages 217--225, 2016.

\bibitem{improvegan}
T.~Salimans, I.~Goodfellow, W.~Zaremba, V.~Cheung, A.~Radford, and X.~Chen.
\newblock Improved techniques for training gans.
\newblock {\em arXiv preprint arXiv:1606.03498}, 2016.

\bibitem{sangkloy2016scribbler}
P.~Sangkloy, J.~Lu, C.~Fang, F.~Yu, and J.~Hays.
\newblock Scribbler: Controlling deep image synthesis with sketch and color.
\newblock {\em arXiv preprint arXiv:1612.00835}, 2016.

\bibitem{schroff2015facenet}
F.~Schroff, D.~Kalenichenko, and J.~Philbin.
\newblock Facenet: A unified embedding for face recognition and clustering.
\newblock In {\em CVPR}, pages 815--823, 2015.

\bibitem{shrivastava2016learning}
A.~Shrivastava, T.~Pfister, O.~Tuzel, J.~Susskind, W.~Wang, and R.~Webb.
\newblock Learning from simulated and unsupervised images through adversarial
  training.
\newblock {\em arXiv preprint arXiv:1612.07828}, 2016.

\bibitem{wang2017zm}
H.~Wang, X.~Liang, H.~Zhang, D.-Y. Yeung, and E.~P. Xing.
\newblock Zm-net: Real-time zero-shot image manipulation network.
\newblock {\em arXiv preprint arXiv:1703.07255}, 2017.

\bibitem{yan2016attribute2image}
X.~Yan, J.~Yang, K.~Sohn, and H.~Lee.
\newblock Attribute2image: Conditional image generation from visual attributes.
\newblock In {\em ECCV}, pages 776--791, 2016.

\bibitem{yi2017dualgan}
Z.~Yi, H.~Zhang, P.~T. Gong, et~al.
\newblock Dualgan: Unsupervised dual learning for image-to-image translation.
\newblock {\em arXiv preprint arXiv:1704.02510}, 2017.

\bibitem{zhu2016generative}
J.-Y. Zhu, P.~Kr{\"a}henb{\"u}hl, E.~Shechtman, and A.~A. Efros.
\newblock Generative visual manipulation on the natural image manifold.
\newblock In {\em ECCV}, pages 597--613, 2016.

\bibitem{zhu2017unpaired}
J.-Y. Zhu, T.~Park, P.~Isola, and A.~A. Efros.
\newblock Unpaired image-to-image translation using cycle-consistent
  adversarial networks.
\newblock {\em arXiv preprint arXiv:1703.10593}, 2017.

\end{thebibliography}
}

	\begin{figure}[!htb]
		\begin{center}
			\includegraphics[scale=0.6]{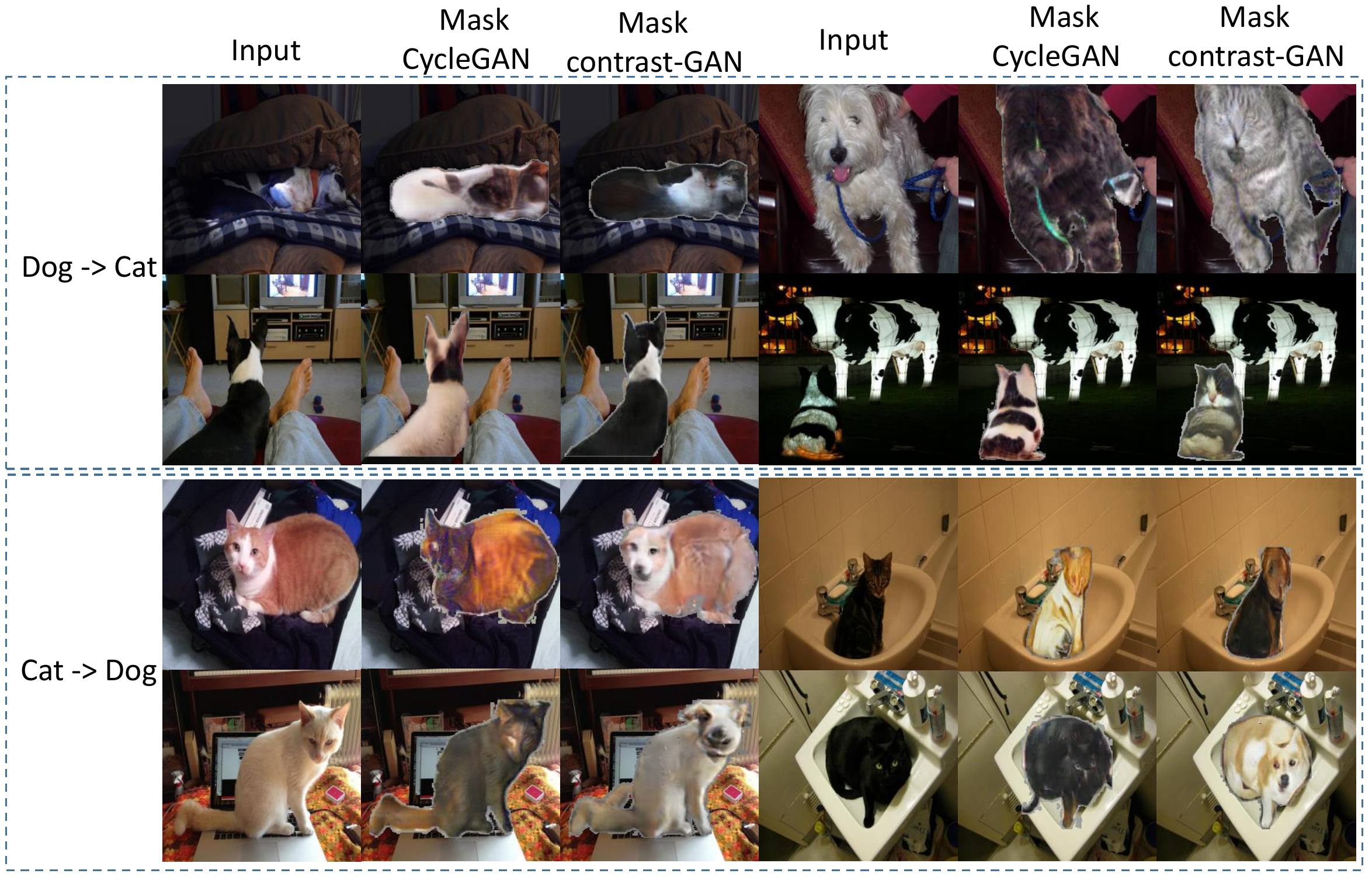}
			\caption{Result comparison between our mask contrast-GAN with mask CycleGAN~\cite{zhu2017unpaired} for translating dog$\rightarrow$cat and cat$\rightarrow$dog on the MSCOCO dataset with provided object masks. It shows the superiority of adversarial contrasting objectiveness over the objectiveness used in CycleGAN~\cite{zhu2017unpaired}.
			} 
			\label{fig:compare_mask_cycle}
		\end{center}
	\end{figure}

	\begin{figure}
		\begin{center}
			\includegraphics[scale=1.1]{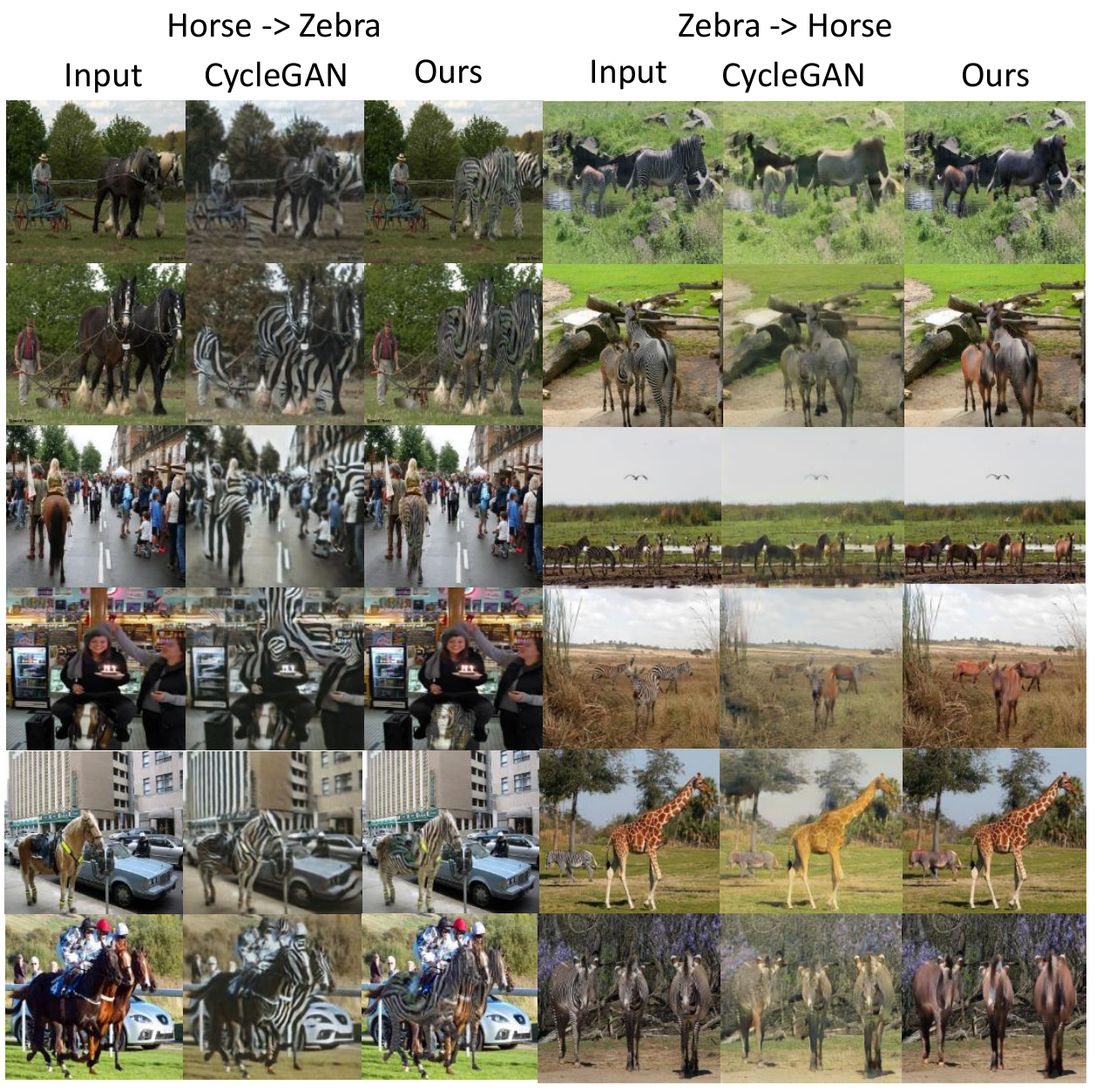}		
			\caption{Result comparisons between our mask contrast-GAN with CycleGAN~\cite{zhu2017unpaired} for translating horse$\rightarrow$zebra and zebra$\rightarrow$horse on the MSCOCO dataset with provided object masks. It shows the effectiveness of incorporating object masks to disentangle image background and object semantics.},  
			\label{fig:compare_mask}
		\end{center}
	\end{figure}

	\begin{figure}[!htb]
		\begin{center}
			\includegraphics[scale=0.43]{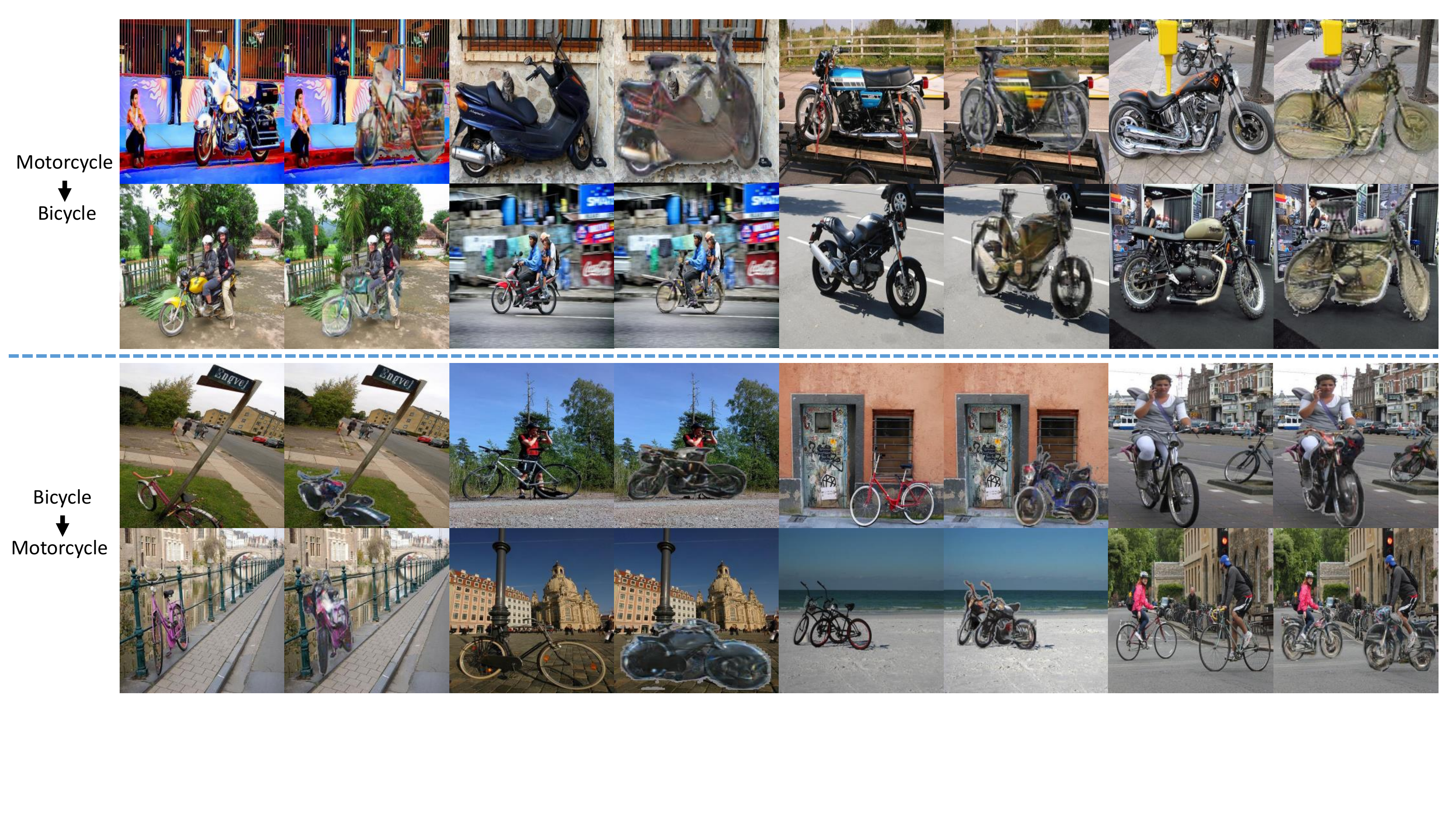}
			\caption{Example results by our mask contrast-GAN for manipulating between bicycle$\leftrightarrow$motorcycle on MSCOCO dataset. For each image pair, we show the original image (left) and manipulated image (right) by specifying a  desirable object semantic.} 
			\label{fig:results}
		\end{center}
	\end{figure}
	
		\begin{figure}[!htb]
		\begin{center}
			\includegraphics[scale=0.53]{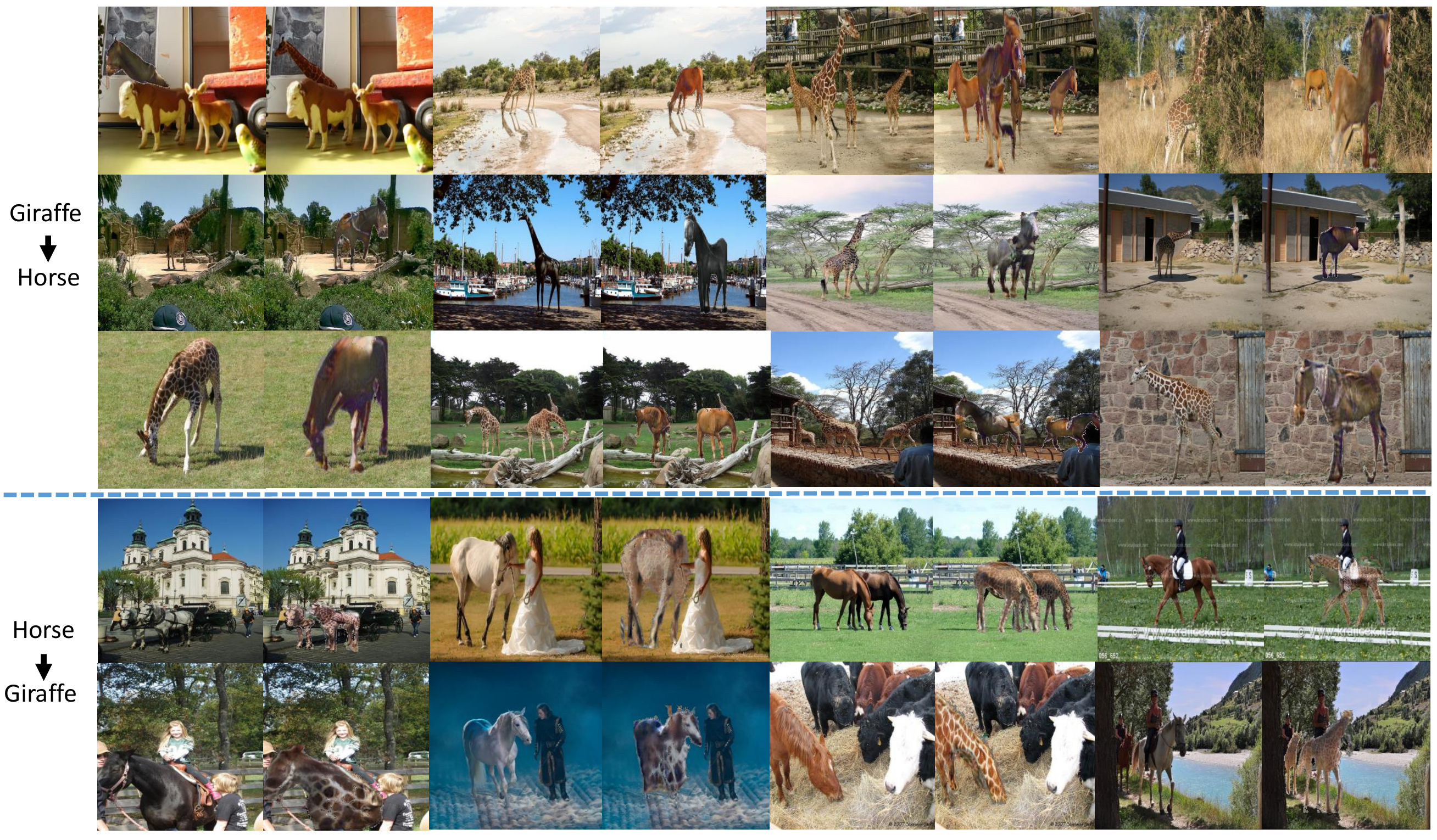}
			\caption{Example results by our mask contrast-GAN for manipulating between giraffe$\leftrightarrow$horse on MSCOCO dataset. For each image pair, we show the original image (left) and manipulated image (right) by specifying a  desirable object semantic.} 
			\label{fig:results}
		\end{center}
	\end{figure}
	
\end{document}